\documentclass{llncs}

\usepackage{times}
\usepackage[dvipdfm]{graphicx}
\usepackage{url}

\newcommand\kihon{atomic action}
\newcommand\kihons{atomic actions}

\newcommand\Kihons{Atomic actions}
\newcommand\decide{action decision making program}

\newcommand\eg{\emph{e.g.}{}}
\newcommand\ie{\emph{i.e.}{}}
\newcommand\QBO{{\tt Q.bo}}
\newcommand\etc{\emph{etc.}{}}

\def\whitedummy#1{%
	\makebox[0pt][l]{
#1
}}

\def\secref#1{Sec.\ref{#1}}
\def\figref#1{Fig.\ref{#1}}

\begin{document}

\title{An Architecture for Autonomously Controlling Robot with
	Embodiment in Real World}

\author{Megumi Fujita\inst{1}\and
	Yuki Goto\inst{2}\and
	Naoyuki Nide\inst{3}\and
	Ken Satoh\inst{4}\and
	Hiroshi Hosobe\inst{5}}
\institute{%
	Graduate School of Humanities and Sciences,
	Nara Women's University, Nara, JAPAN
	\email{{\tt sabo\whitedummy{kuma}ten@ics.nara-wu.ac.jp}}\and
	Research Institute for Mathematical Sciences,
	Kyoto University, Kyoto, JAPAN\and
	Faculty, Division of Natural Sciences,
	Nara Women's University, Nara, JAPAN\and
	Principles of Informatics Research Division,
	National Institute of Informatics, Tokyo, JAPAN\and
	Faculty of Computer and Information Sciences,
	Hosei University, Tokyo, JAPAN%
}





\maketitle

\begin{abstract} 
In the real world,
robots with embodiment face various issues such as
dynamic continuous changes of the environment
and input/output disturbances.
The key to solving these issues can be found in daily life;
people `do actions associated with sensing'
and `dynamically change their plans when necessary'.
We propose the use of a new concept,
enabling robots to do these two things,
for autonomously controlling mobile robots.
We implemented our concept to make two experiments
under static/dynamic environments.
The results of these experiments show that
our idea provides a way to adapt to
dynamic changes of the environment in the real world.
\end{abstract}



\section{Introduction}
We aim to make
robots who decide their actions
to achieve their goals in the real world.

It is known that
robots in the real world
are exposed to various issues
that are never present in the
virtual or formally modeled worlds \cite{Pfeifer06}.
One of them is that, as is widely recognized,
the real world is highly dynamic;
there are continuous changes of circumstances.
Another is that
the physical devices
can never escape from
input/output disturbances,\label{io_disturb}
which cause failures in robots' actions.
These issues arise from the fact that robots have embodiment in the real world. 
\label{real_world_fluctuation}

The key idea to solving these issues can be discovered
in daily life.\label{daily_life}
When we walk,
we are sensing various things, \eg{} traffic signals, the roaring of a car engine,
the presence of vending machines,
while moving ahead.
We may stop moving since the traffic signal is red,
or reflexively jump back since we sense danger from the roaring.
Here, sensing and action work together rather than being separated,
and the plan is updated dynamically.
We use such actions, \eg{} `walking forward until finding some obstacle',
to construct plans to achieve our goals.
The robots who have embodiment also need to have
such ability to act in the real world.

However, traditional planning theory separates sensing and actions.
Since the origin of a conceptual model for planning \cite{malik2004} is
the model of state-transition systems (also called discrete-event
systems), it equates an \kihon{} with a step in a state machine. In
other words, \kihons{} are considered to be carried out in a
moment, and no sensing takes place during that action.
As discussed above, this feature is not suitable for implementing
such behaviors.

\begin{figure}
\centering
\includegraphics[width=8cm]{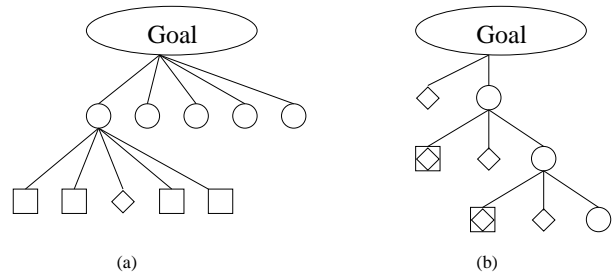}
\caption{Circle denotes a subgoal, box means an \kihon{}, and
diamond denotes sensing (one kind of \kihons{}).
The box with a diamond denotes the \kihon{} with sensing.
Here, (a) is a diagram of general planning theory,
and (b) is that of our proposal.}
\label{proposal}
\end{figure}

We propose a way to implement the key idea described above.
To do this, we associate actions with sensing in implementation
rather than separating them,
and instead let the next action be selected dynamically.
In our system,
\Kihons{} are implemented with sensing and storing 
external perceptions. This is the main difference between general
automated planning and ours. (\figref{proposal})

The \decide{}, which selects suitable \kihons{},
is implemented by Prolog.
It takes the robot's goal as a Prolog goal,
and decomposes it to the next action and the remaining subgoal;
\ie{} each step of the derivation procedure of Prolog and
the robot's each action are interconnected.
We also describe our experiment on motion planning in the real world.

This paper is organized as follows:
\secref{planning_in_the_realworld} explains
the difficulty of making motion planning with dynamic planners
and describes how our research has overcome it.
\secref{robot_detail} describes in detail the robot \QBO{}
which we used in the experiment. The following section
\secref{motion_planning} shows aspects of our computer programs, and how
they work. In \secref{experiment}, we show our experiments in the
real world. Related works are shown in
\secref{related_works}. Finally,
in \secref{conclusion}, we provide some discussions
on the remaining issues and conclude.

\section{Motion planning in real world}
\label{planning_in_the_realworld}
Many ``dynamic planners'' have been published,
which respond to dynamic changes in the environment.

For example, in \cite{HOTRiDE}, Ayan et al.\ use it for
Noncombatant Evacuation Operations planning (\eg{} selecting the means of
transportation and the route to be followed),
and in \cite{Therapy}, S\'{a}nchez-Garz\'{o}n et al.\ use it for
therapy planning systems in a real clinical environment. 

However, traditional dynamic planners in the real world
cannot avoid
making a significant assumption that
the world can be accurately described as
a state-transition system based on atomic actions.
In other words,
it must be assumed that
the effects, pre- and post-conditions of atomic actions
can be accurately described.
However, due to the I/O disturbances of devices described in
\secref{io_disturb},
for example, `to go ahead exactly with 1 meter toward north'
is very hard for robots in the real world.
If a robot tried to do so and actually proceeded by 0.98 meter,
a dynamic planner would not consider that
the robot has normally reached the expected state,
and it will consider that situation as a failure of the plan.

To overcome this difficulty,
we have taken a different approach.
We implement atomic actions analogous to
our natural behaviors in daily life discussed in \secref{daily_life},
as follows.

\begin{itemize}
\item Sensing and actions working together \\
Instead of the `sensing after action' manner
used by traditional dynamic planners,
we combine sensing and action,
just like we sense surroundings while we walk.
In our method,
an atomic action is done while always receiving
data from the sensor,
and it terminates when
some terminating condition (\eg{} a certain period of time
has been passed, an obstacle is found, \etc{}) is satisfied.
At that time, information about the current environment
of the robot will be sent to the \decide{} which is used to
determine the next action.
Any atomic action is guaranteed to terminate eventually\label{notimeout}
(or, to be strict, we design atomic actions to satisfy this property).
\item Dynamic action selection depending on
	current condition \label{interleave}\\
Using the information sent from the sense-and-action part described above,
the \decide{} decides the best suited atomic action to be executed next.
It uses rules written in Prolog
and always returns some atomic action (see \secref{action_decision}).
It is very much like
teleoreactive logic programs \cite{Langley06} (but see
\secref{related_works}).
\end{itemize}

In \secref{experiment}, by our experiments,
we show that
even if the robot's actions are not sufficiently accurate\label{nnaccurate},
it can continue to act
and eventually reach its goal.
In the experiments,
the robot sometimes lost sight of the target
due to the inexactness of the object recognition routine
implemented in the SVM (Support Vector Machine) \cite{Bishop07}.
However, while going toward the approximate direction of the target,
it found the target again and went toward it.
We also show that,
under both the static environment and the dynamic one,
the robot reached the target using
the same pair of \decide{} and atomic action set.

\section{Detailed description of robot}
\label{robot_detail}
\begin{figure}
\centering
\includegraphics[width=4.4cm]{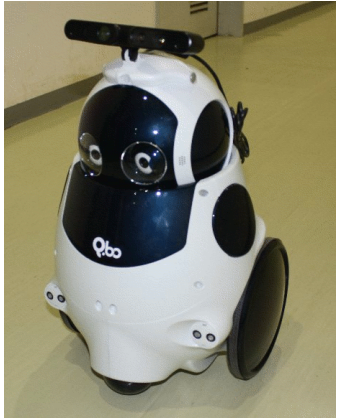}
\caption{Q.bo Lite Evo}
\label{QBO}
\end{figure}

We used the robot `\QBO{} Lite Evo'. A Spanish company, TheCorpora S.L.,
sells this robot, and distributes {\tt Q.bo}'s particular Linux distribution
based on Ubuntu. We illustrate \QBO{} in \figref{QBO}.

\QBO{} moves with 2 side wheels (rear), and 1 caster wheel
(front). There are many applications, \eg{} face recognition, speech
recognition, and object recognition, and so on. We can control \QBO{}
with a robotic software platform, Robot Operating System (ROS) \cite{ROS09}.

\section{Motion planning}
\label{motion_planning}
For motion planning, we write the \decide{} and computer programs that
run \QBO's \kihons{}. \Kihons{} are described with Python, and the
\decide{} is implemented with SWI-Prolog \cite{SWI11}.

In our system,
thinking and action are generally run by turn. At first
\QBO{} runs \decide{} to get a new \kihon{}. Second, \QBO{} runs this
\kihon{} while sensing in the real world, and passes the requisite
information on \decide{} after the \kihon{} terminates. \QBO{} continues
these two processes.

\subsection{\Kihons}
We control \QBO{} with ROS, which uses ROS Topic for sharing information.
In this experiment, we use ROS Topics with a focus on motion
planning. The requisite information (\eg{} start a motor to move,
recognize objects, measure distance, direction, and velocity, \etc) are
shared using ROS Topics.

\Kihons{} are made using these topics, which are updated in real
time. In this way, \QBO{} can move
in response to changes in their environment
in a way that is similar to reflex actions.

A detailed explanation of \kihons{} follows%
\footnote{As mentioned in \secref{notimeout}, any atomic action terminates
eventually, \ie{} no action has a timeout. In particular,
{\tt search\_Qbo} currently assumes that
it can eventually find a direction without an obstacle.}.
\begin{itemize}
\item \verb|looking_Qbo|
\begin{itemize}
\item \QBO{} looks for the target 
while moving his head.
It returns the direction of the target, when he detects it.
\item If he cannot detect the target, it does nothing and terminates
this process.
\end{itemize}
\item \verb|search_Qbo(Direction)|
\begin{itemize}
\item \QBO{} turns around, searches for a direction, in which
there are no obstacles, and \QBO{} moves forward in this direction.
When \QBO{} searches for such a direction, he makes an effort to
choose the direction as close to the argument \verb|Direction| as possible.
\end{itemize}
\item \verb|forward_Qbo(Direction)|
\begin{itemize}
\item \QBO{} moves ahead for a fixed distance until some obstacle is found.
\item If \QBO{} finds some obstacle when he is moving forward, stops moving,
and terminates this process.
\end{itemize}
\end{itemize}

In \verb|looking_Qbo|, object recognition for finding the target
is currently implemented using libSVM \cite{Chang11}.
It was trained using about 40 images of the target and 40
images of the different things, and classifies the images from
\QBO's camera into ones of the target and of the different things.
Currently the accuracy of the classification is not so high.
We are planning to compare this with one implemented using OpenCV.

\subsection{\decide{}}\label{action_decision}
\QBO{} runs \decide{} to get a new \kihon{}
that is suitable for the present
circumstance. As shown in \secref{motion_planning}, \decide{} is
implemented with Prolog. The derivation of a Prolog enables \QBO{} to
infer a new \kihon{}, and the unification in a Prolog enables \QBO{} to
update requisite information such as sensor information.

By and large, two kinds of rules make up \decide{}. One group relate to
storing sensor information, the other to conditional execution of
\kihon{}.

Abstract descriptions of the \decide{} are as the following
pseudo-Prolog code.

\begin{quote}
\begin{verbatim}
Toplevel :- Initialize, Goal.
Goal :- Percept, Goal'.
Goal' :- Condition, !, Atomic_action, Goal.
\end{verbatim}
\end{quote}

The following is a detailed explanation of \decide{}.

\begin{enumerate}
\item initialize \\
Initialize information from stored sensor information of \QBO{}.\\ 
In particular, \verb|Initial_state/3| is updated to record \QBO's
initial direction and location.
\item perception \\
\QBO{} runs some specific \kihon{} (\eg{} \verb|looking_Qbo|) to
have external perceptions to get some information
like its position and direction.
\item branch condition \\
The branch, which is the first to have its condition satisfied with
the perceptions, is chosen to execute.
\item run a \kihon{} \\
\QBO{} runs following \kihon{}. At this point,
\QBO{} waits for a termination of the \kihon{}.
\end{enumerate}

The \decide{} continues to repeat from 2 to 4.
We show almost the whole \decide{} code below.

\begin{quote}
\begin{verbatim}
/* Action decision making routine. First rule of search_target/7
        is selected at only the first time that initial state
        information was obtained and returns no_operation action. */
search_target(D,X,Y,Op,Obj,Input,Output) :-
    retract(first),
    assert(initial_state(D,X,Y)),
    Op = none.

/* The main rule for Qbo's action decision making. */
search_target(D,X,Y,Op,Obj,Input,Output) :-
    get_directions(D,I),
    around_search(F,FD,Input,Output),
    decide_action(F,FD,D,I,Op,Obj).

/* Get the directions which Qbo towards at the initial state. */
get_directions(Direction,Initial_Direction) :-
    initial_state(Initial_Direction,_,_).

/* Send the command "looking_Qbo" to Qbo for looking around
        and search the target. */
around_search(Found, Found_Direction,Input,Output) :-
    write(Output,looking_Qbo),nl(Output),flush_output(Output),
    recognize_target(Found, Found_Direction, Input).

/* If Qbo recognizes the target, Qbo sends back "True" message
        and the argument 'Found' is unified to it. Otherwise,
        'Found' is unified to "False". */
recognize_target(Found,Found_Direction,Input) :-
    read_line_to_codes(Input, T1, T2),
    name(Found, Found_Direction, T1, T2).

/* If Qbo recognizes the target, Qbo tries to go toward the
        target. */
decide_action(F,FD,D,I,Op,Obj) :-
    F = true, !,
    write('target found'), nl,
    go_forward(D,FD,Op,Obj).
/* Otherwise, Qbo goes toward the initial direction. */
decide_action(F,FD,D,I,Op,Obj) :- !,
    write('target not found'), nl,
    go_forward(D,I,Op,Obj).

/* The first rule of go_forward/4 sends the command which
        leads Qbo to go forward avoiding the obstacle. */
go_forward(Direction, Initial_Direction, Operator, Obj) :-
    Obj = 1, !,
    Operator = search_Qbo(Initial_Direction).
/* The second one sends the command to go forward by a fixed
        distance until Qbo finds the obstacle. */
go_forward(Direction, Initial_Direction, Operator, Obj) :- !,
    Operator = forward_Qbo(Initial_Direction).

/* Top level routine (which acts as a TCP client); creates
        socket for connecting to the program (a TCP server)
        managing Qbo's atomic action, and calls the robot's
        goal (start searching for the target). */
client(H, P):-
    tcp_socket(S),
    tcp_connect(S, H:P, I, O),
    prompt(_, ''),
    searching(S, I, O).

/* Robot's goal; Searching the target. */
searching(S, I, O) :-
    /* Read the sensor information from Qbo */
    read_line_to_codes(I, T1,T2,T3,T4),
    name(D,X,Y,Obj,T1,T2,T3,T4),

    /* Call the action decision making routine. */
    search_target(D,X,Y,Operator,Obj,I,O),
    /* Send the atomic action Operator to Qbo. */
    write(O, Operator), nl(O),
    flush_output(O),
    /* Recursively call this goal. */
    searching(S, I, O).
\end{verbatim}
\end{quote}

\section{Experiments in real world}
\label{experiment}
We carried out the following two experiments.

\begin{itemize}
\item {\it First experiment} \\
We placed \QBO{}, an obstacle, and the target in the space for our
experiments. The target is put ahead of \QBO{}, but the obstacle
obstructs the way to the target. \QBO's goal is to reach the target
while avoiding the obstacle.
\item {\it Second experiment} \\
We placed \QBO{}, and the target in the same space. The target is put
ahead of \QBO{} again, and there are no obstacles on the way to the
target at the beginning. When \QBO{} finds the object, we set the
obstacle on the way to the target. \QBO's goal is to reach the
target, avoiding the obstacle. (This is equivalent to the prior goal.)
\end{itemize}

The first experiment shows how our programs work under the static
environment. The second one shows how to get over
dynamic changes of the environment in the real world. In these two
experiments, we use the same \decide{} and \kihons{} because the goal of
these is the same, \ie{} ``reach the target''.

\subsection{Experiment results}

\begin{figure*}[tbp]
\def\figwidth{.33\textwidth}\def\figsep{7pt}%
\centering
\makebox[0pt]{%
\begin{tabular}{ccc}
\includegraphics[width=\figwidth]{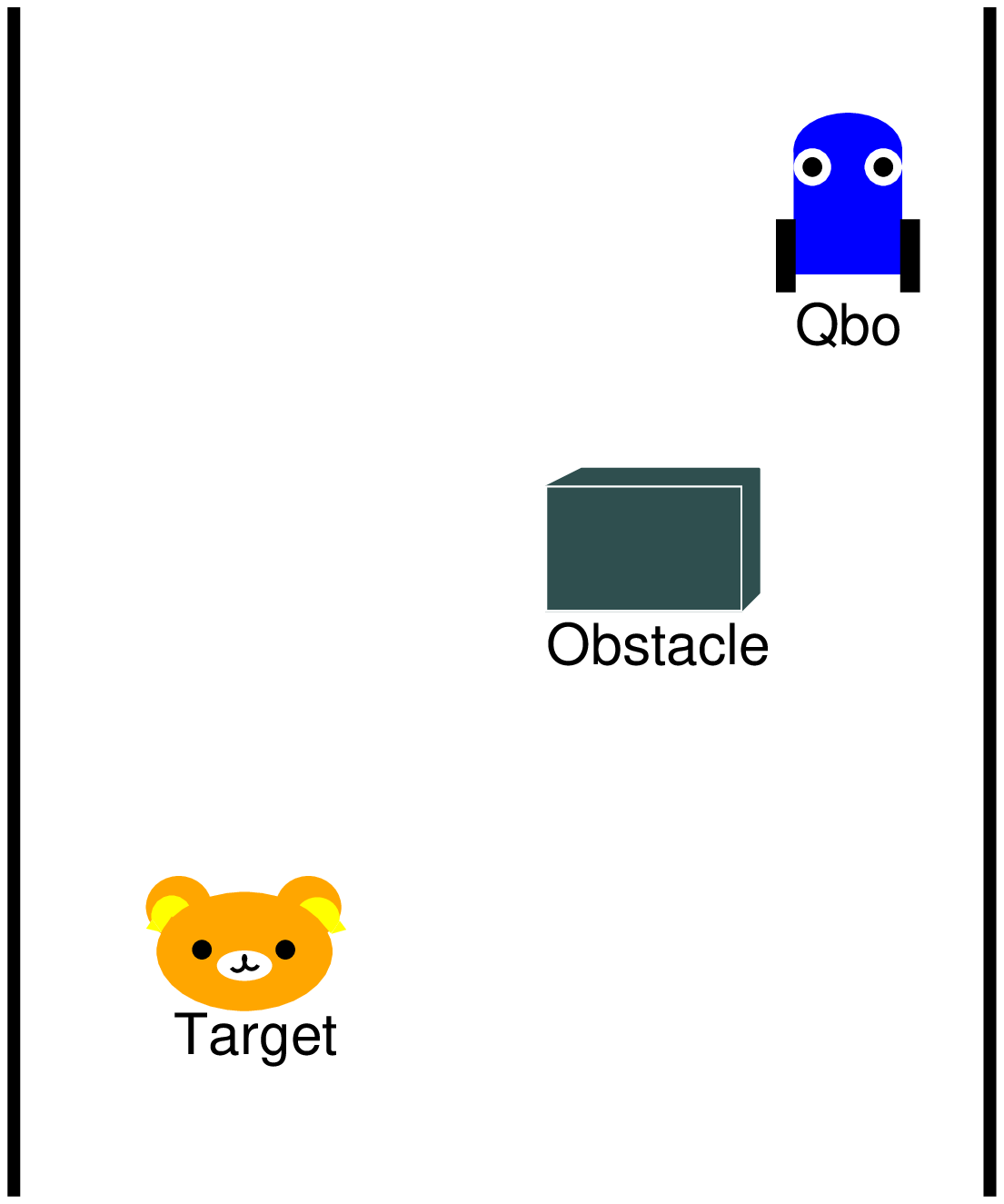}&
\includegraphics[width=\figwidth]{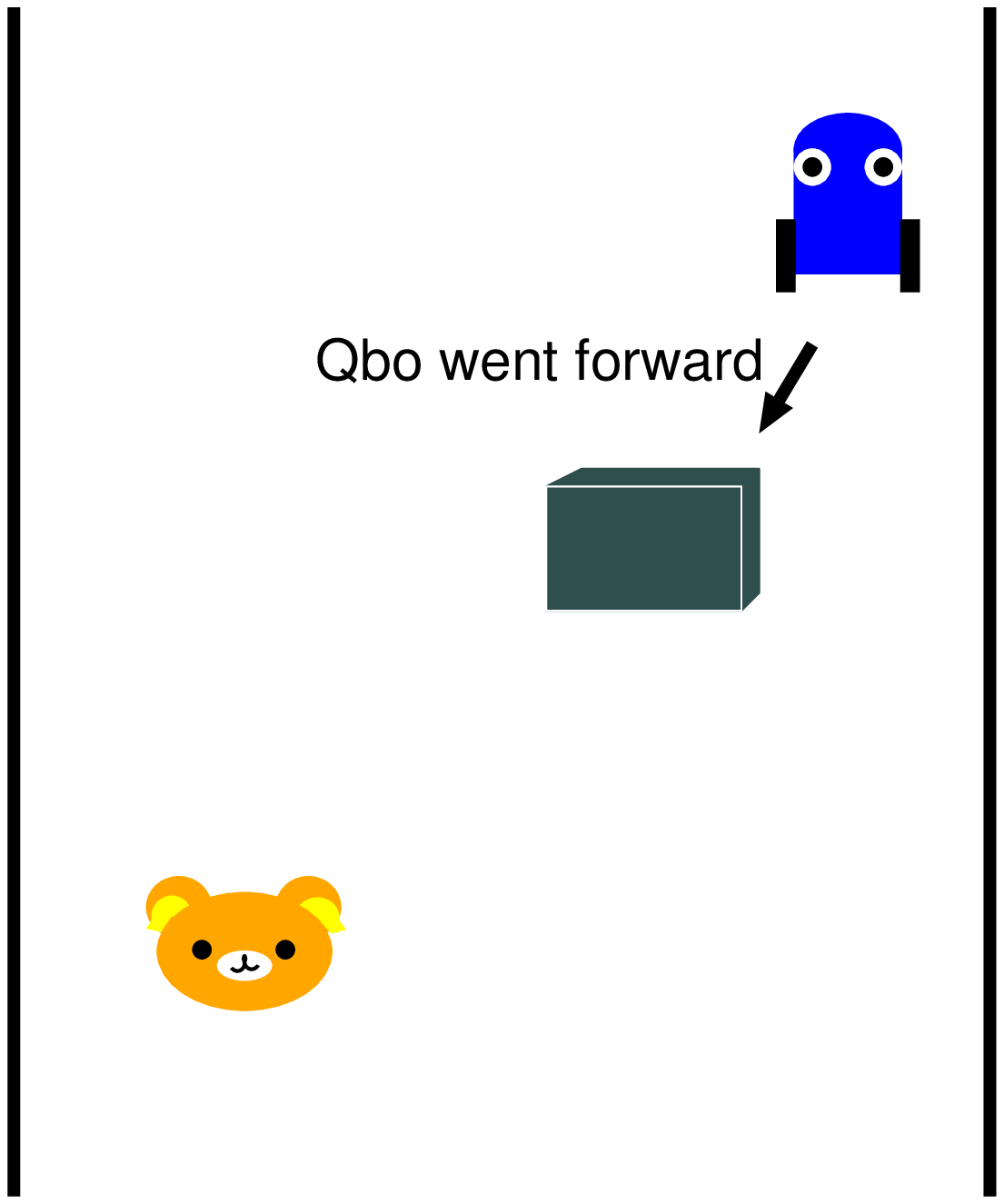}&
\includegraphics[width=\figwidth]{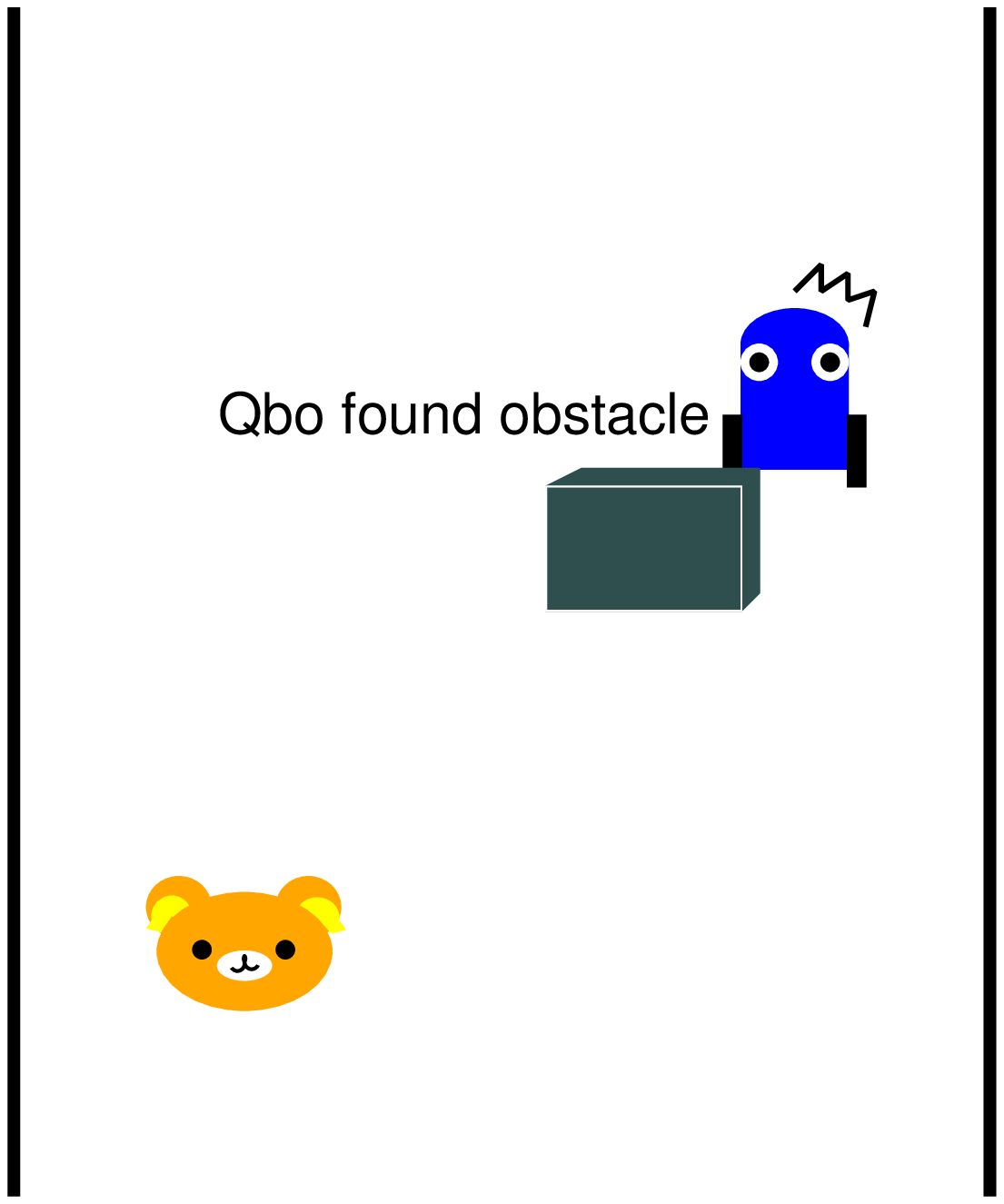}\\
(1) Initial state.&(2) \QBO{} found target and&(3) \QBO{} found obstacle.\\
		&went forward to it. &			\\[\figsep]
\includegraphics[width=\figwidth]{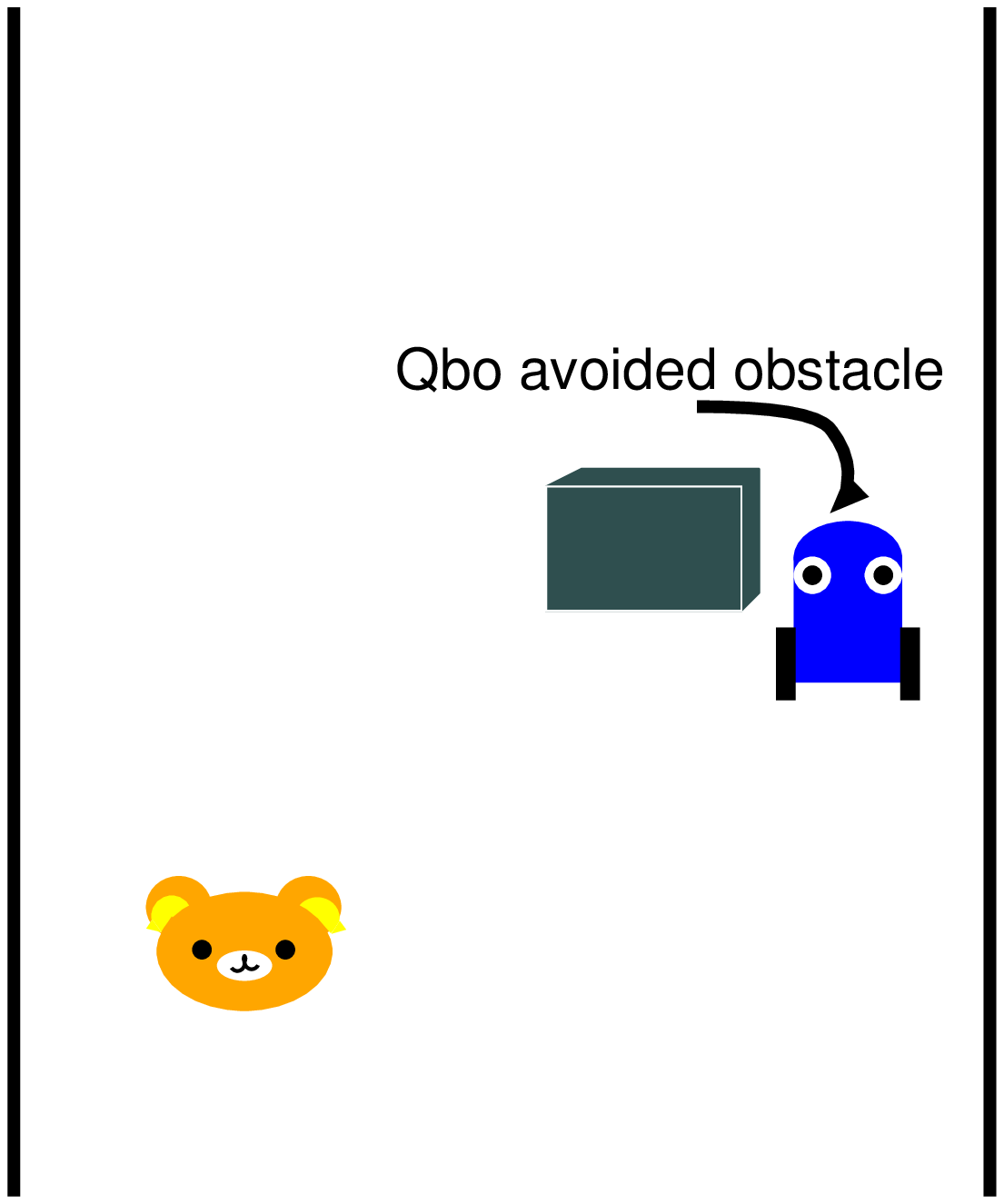}&
\includegraphics[width=\figwidth]{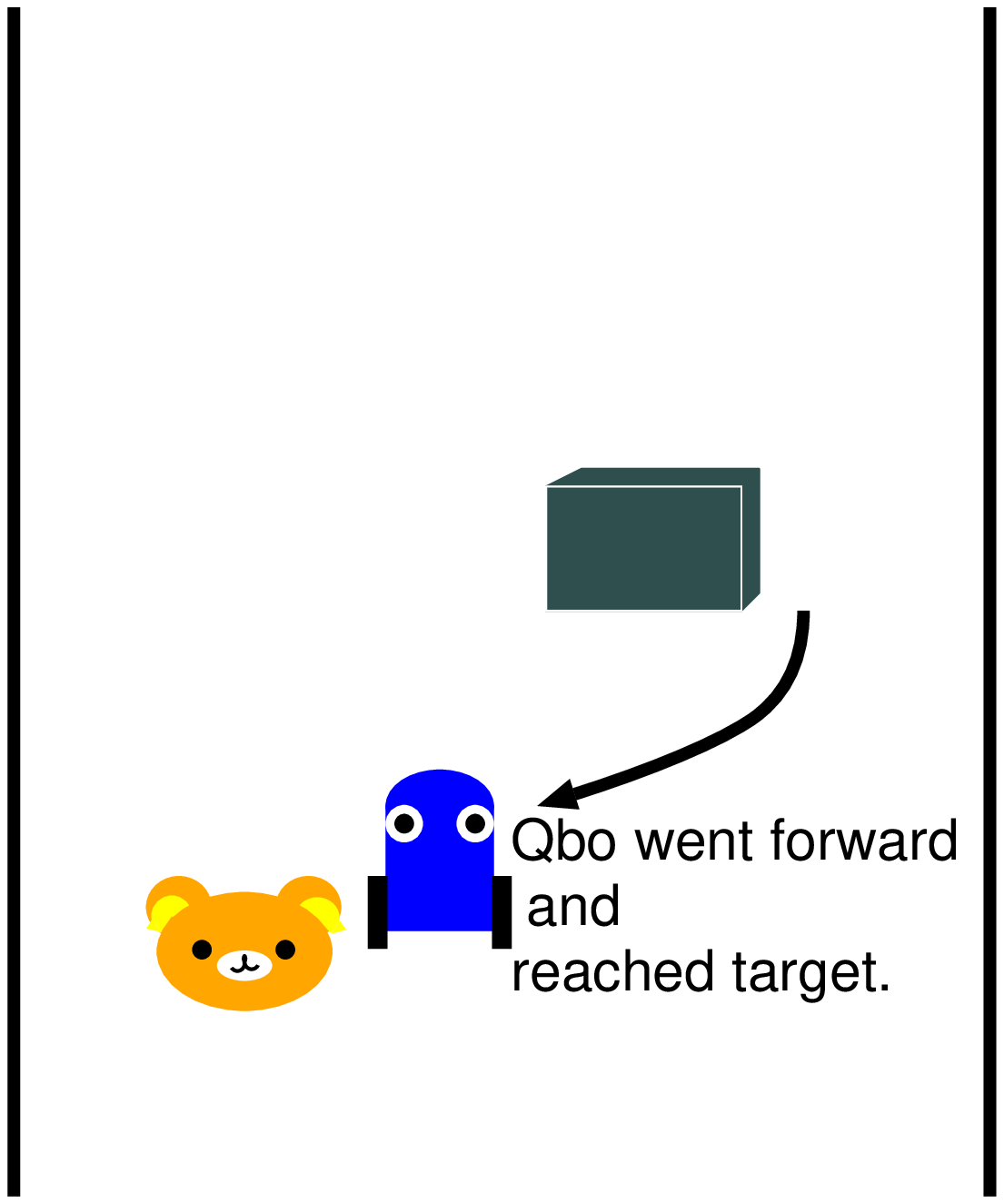}&\\
(4) \QBO{} avoided obstacle.&(5) \QBO{} reached a point&\\
			&in front of target.&
\end{tabular}%
}
\caption{First experiment}
\label{Fig:experiment1}
\end{figure*}

\begin{figure*}[!p]
\centering
\vskip-5pt
\scalebox{1}[.9]{%
	\includegraphics[width=0.93\textwidth]{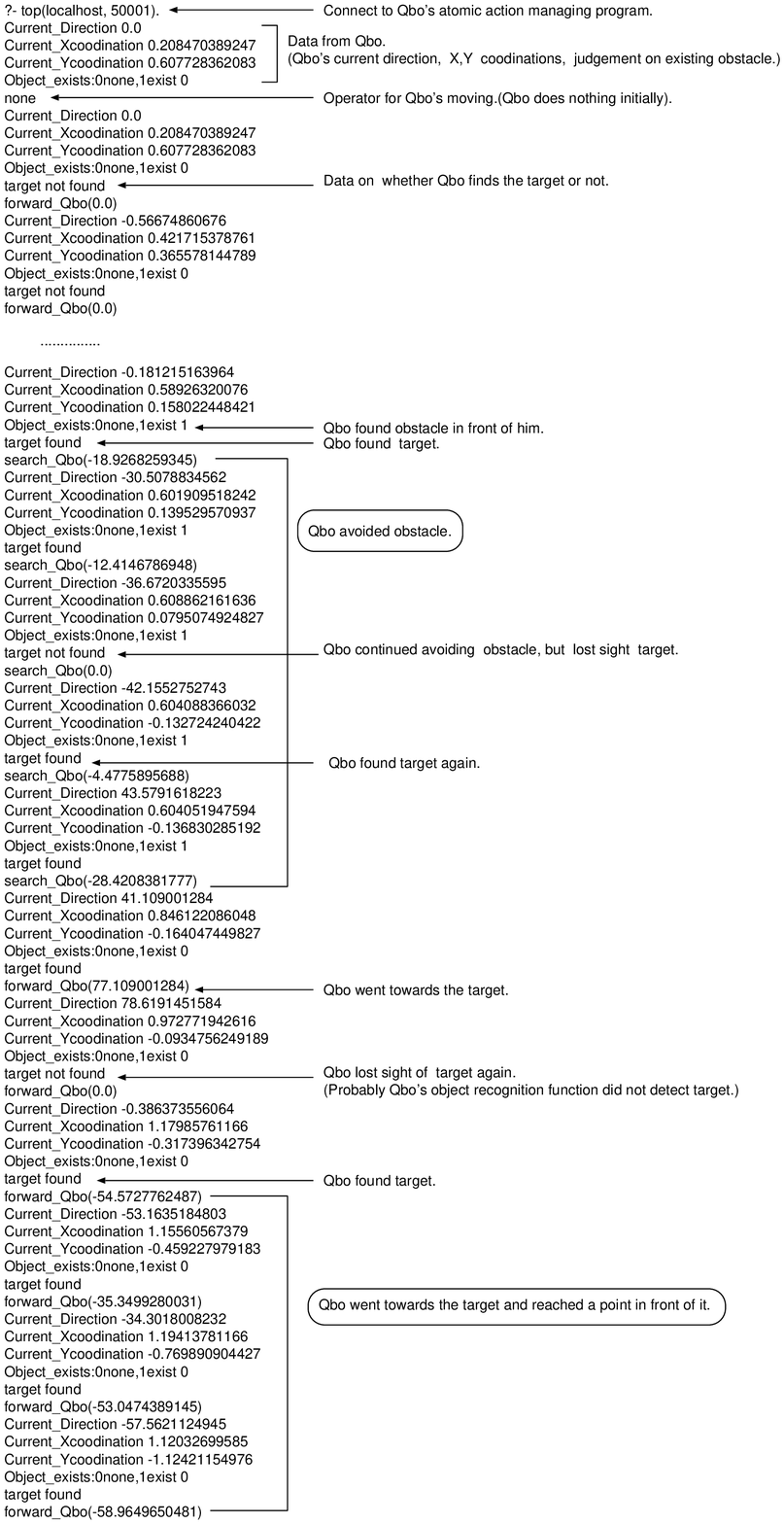}}
\vskip-5pt
\caption{log messages of first experiment result}
\label{Fig:log}
\end{figure*}

\begin{figure*}[tbp]
\def\figwidth{.48\textwidth}\def\figsep{7pt}%
\centering
\makebox[0pt]{%
\begin{tabular}{cc}
\includegraphics[width=\figwidth]{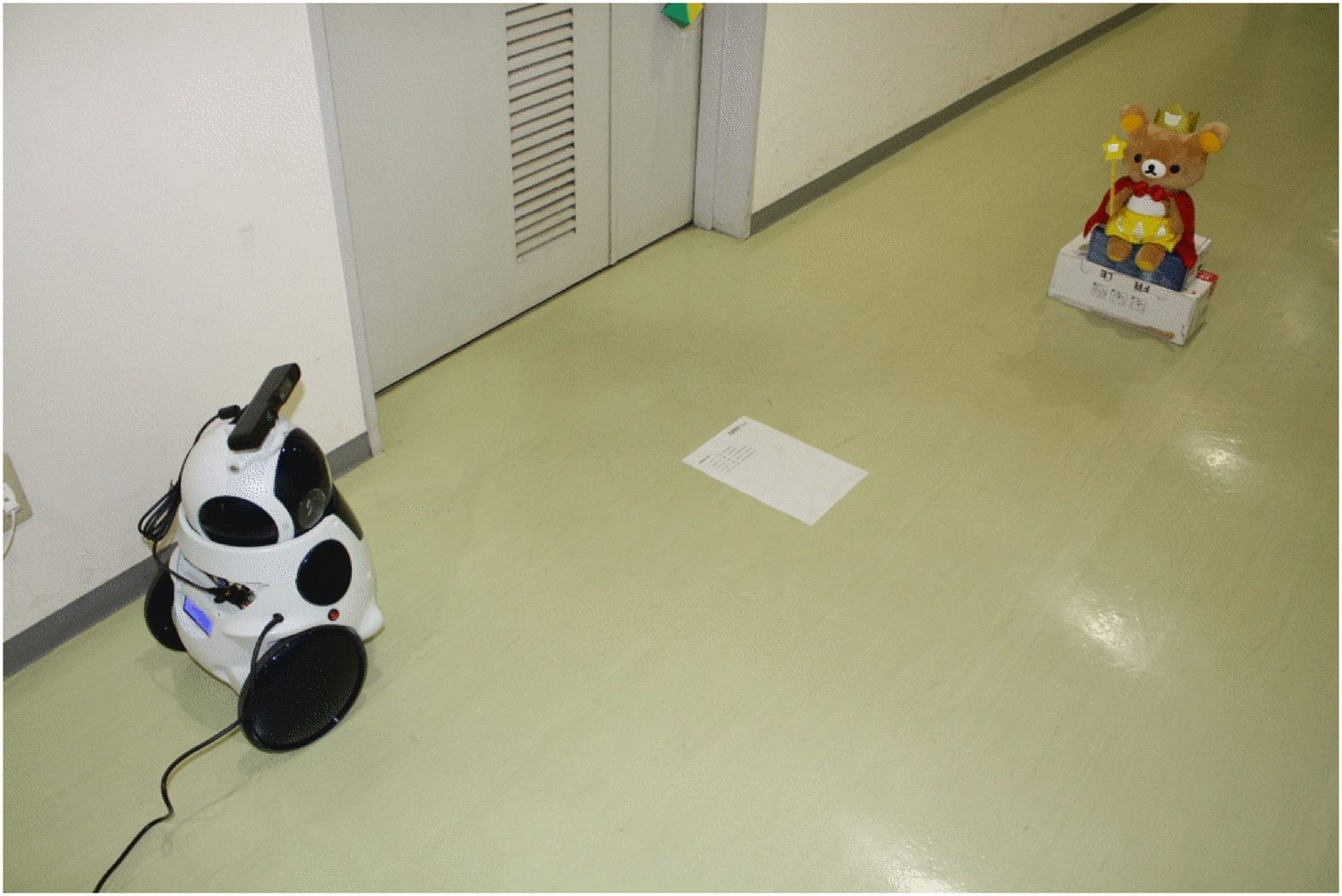}&
\includegraphics[width=\figwidth]{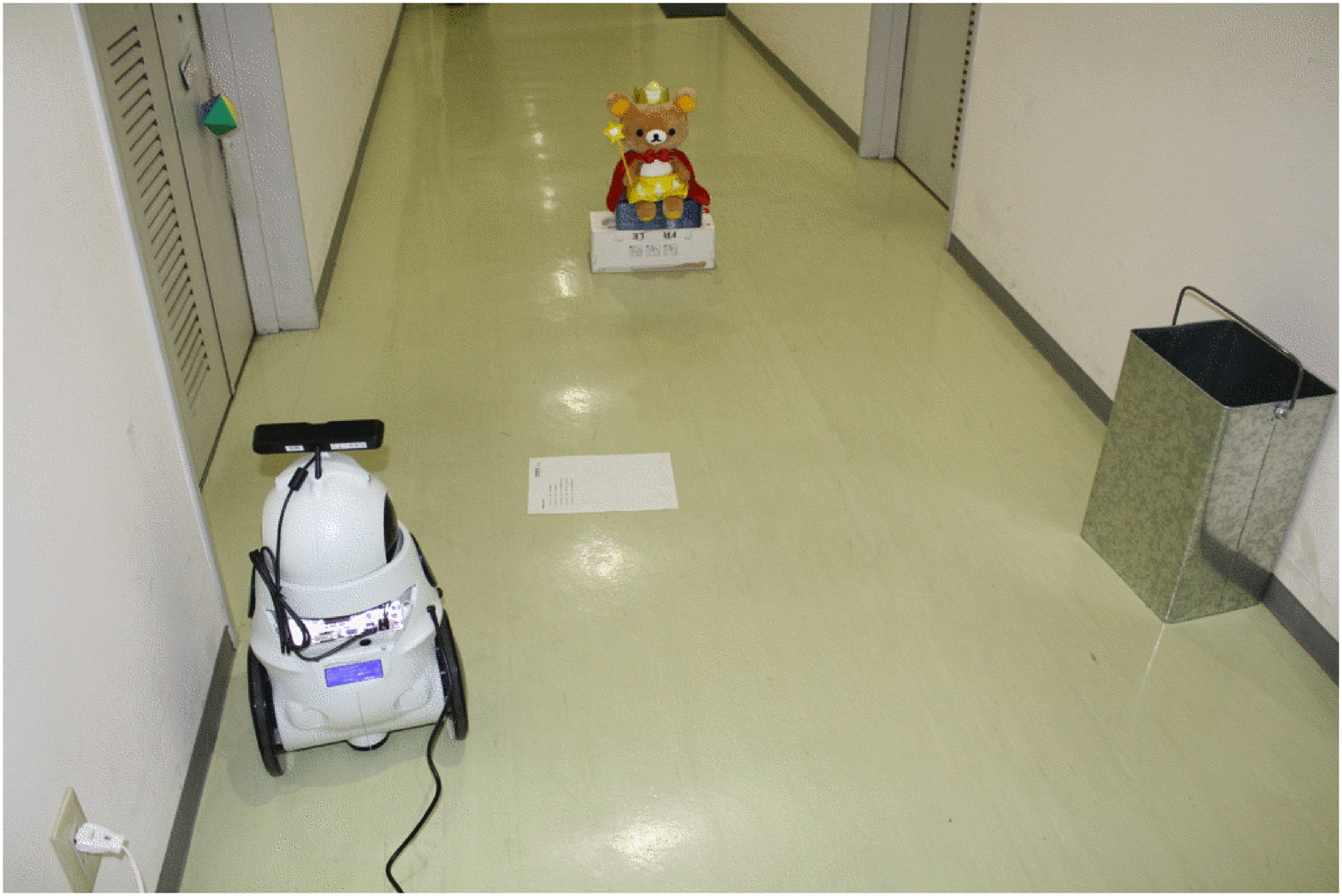}\\
(1) Initial state. &(2) \QBO{} found target
		and went towards it.\\[\figsep]
\includegraphics[width=\figwidth]{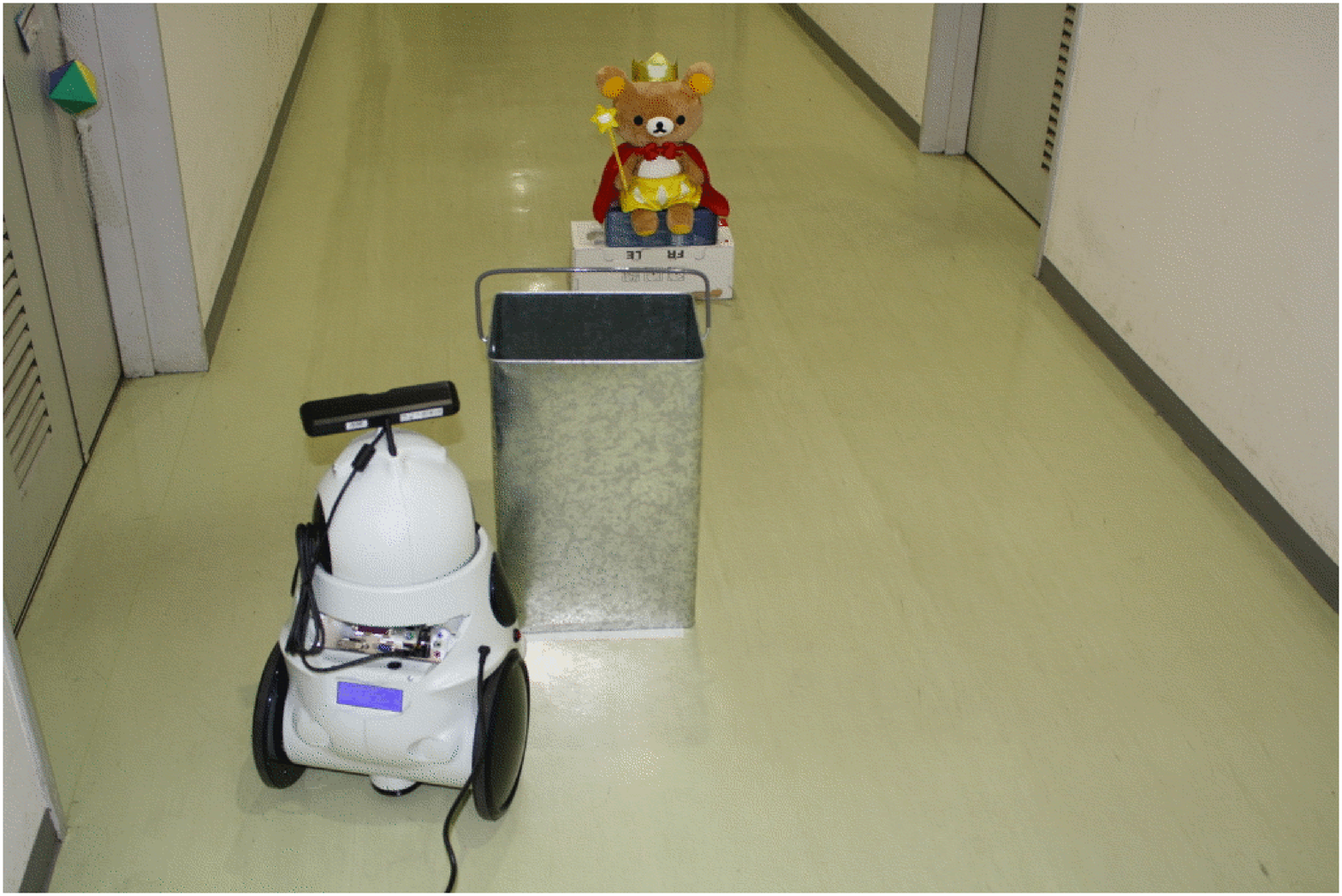}&
\includegraphics[width=\figwidth]{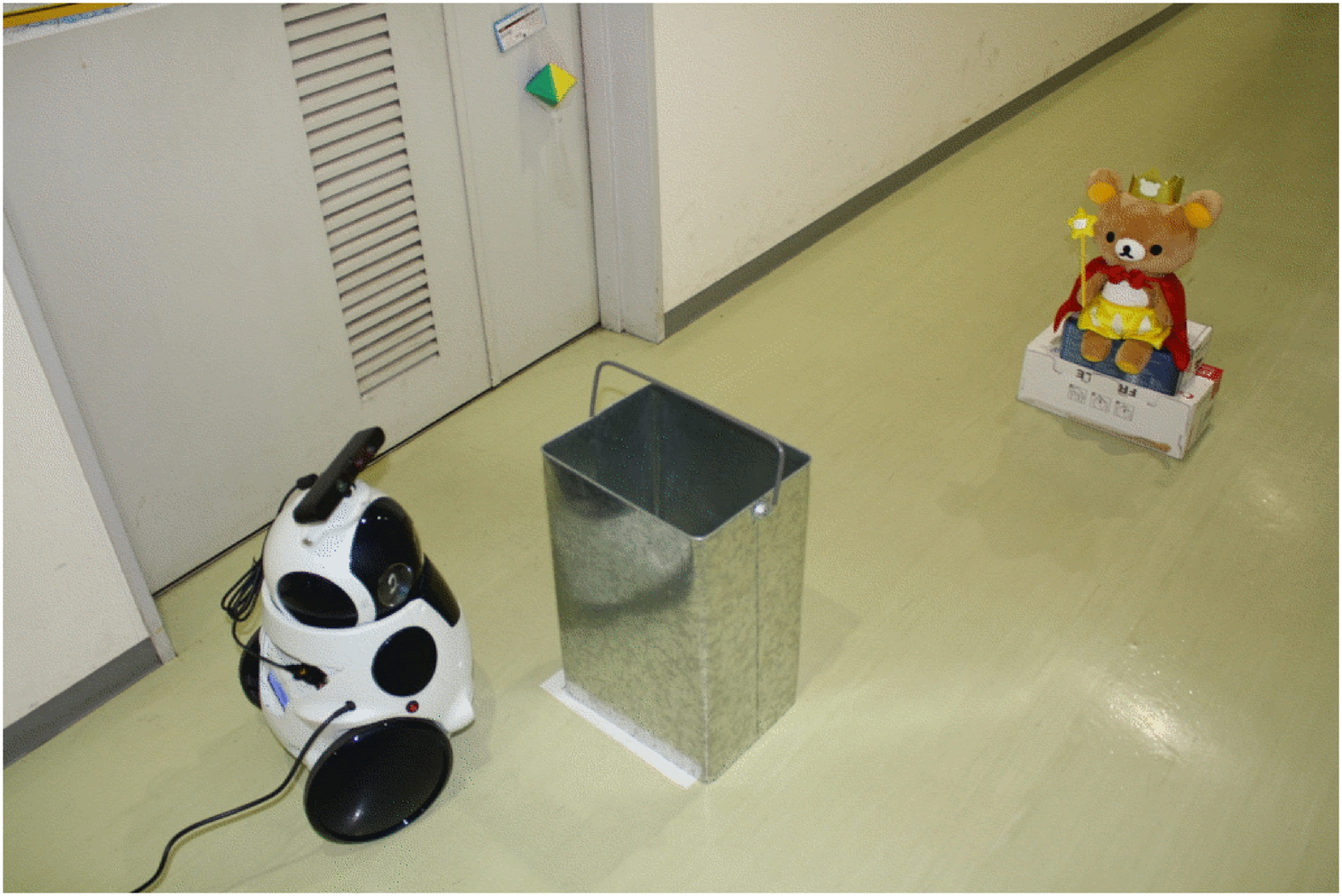}\\
(3) \QBO{} found obstacle.&(4) \QBO{} could not see target.\\[\figsep]
\includegraphics[width=\figwidth]{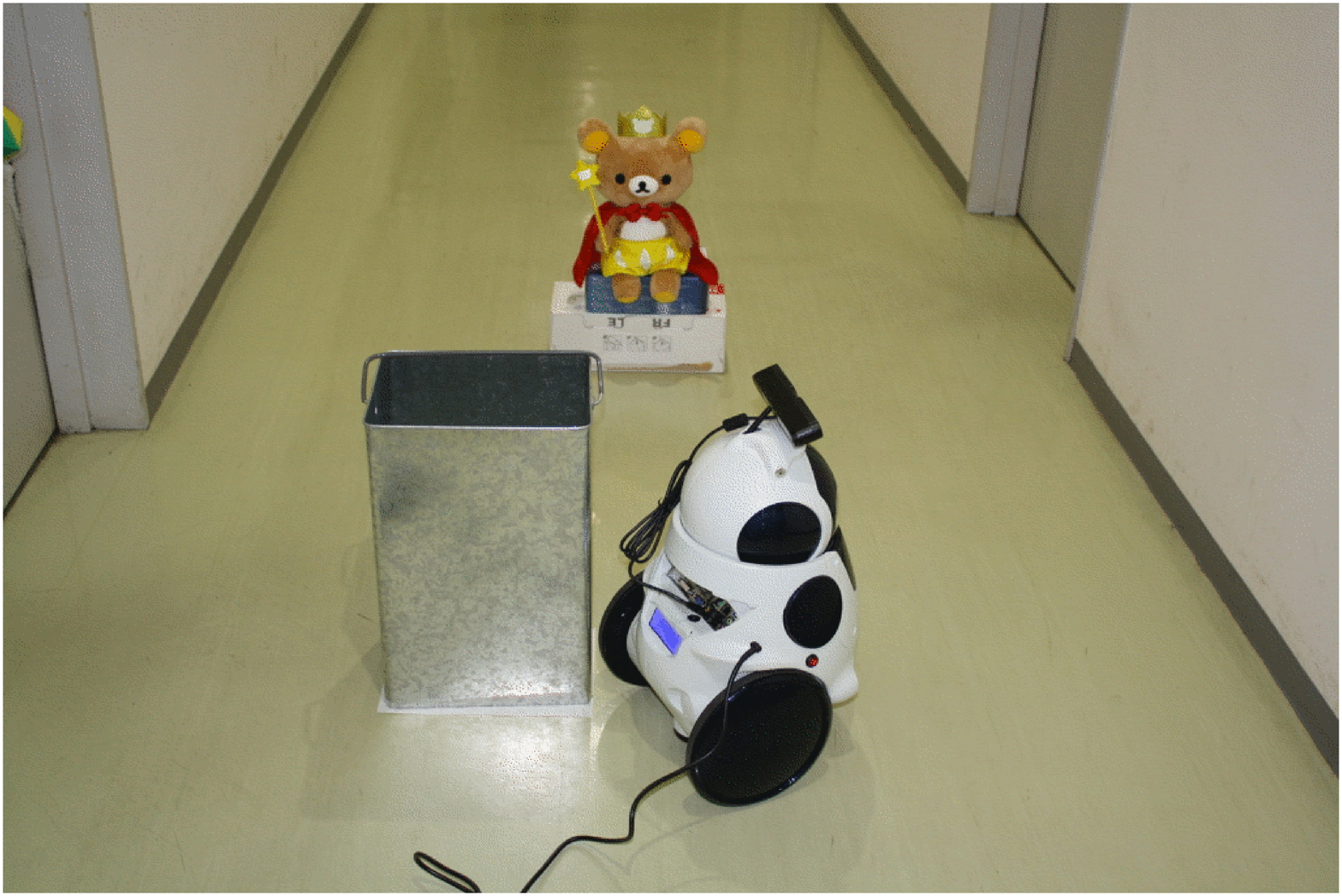}&
\includegraphics[width=\figwidth]{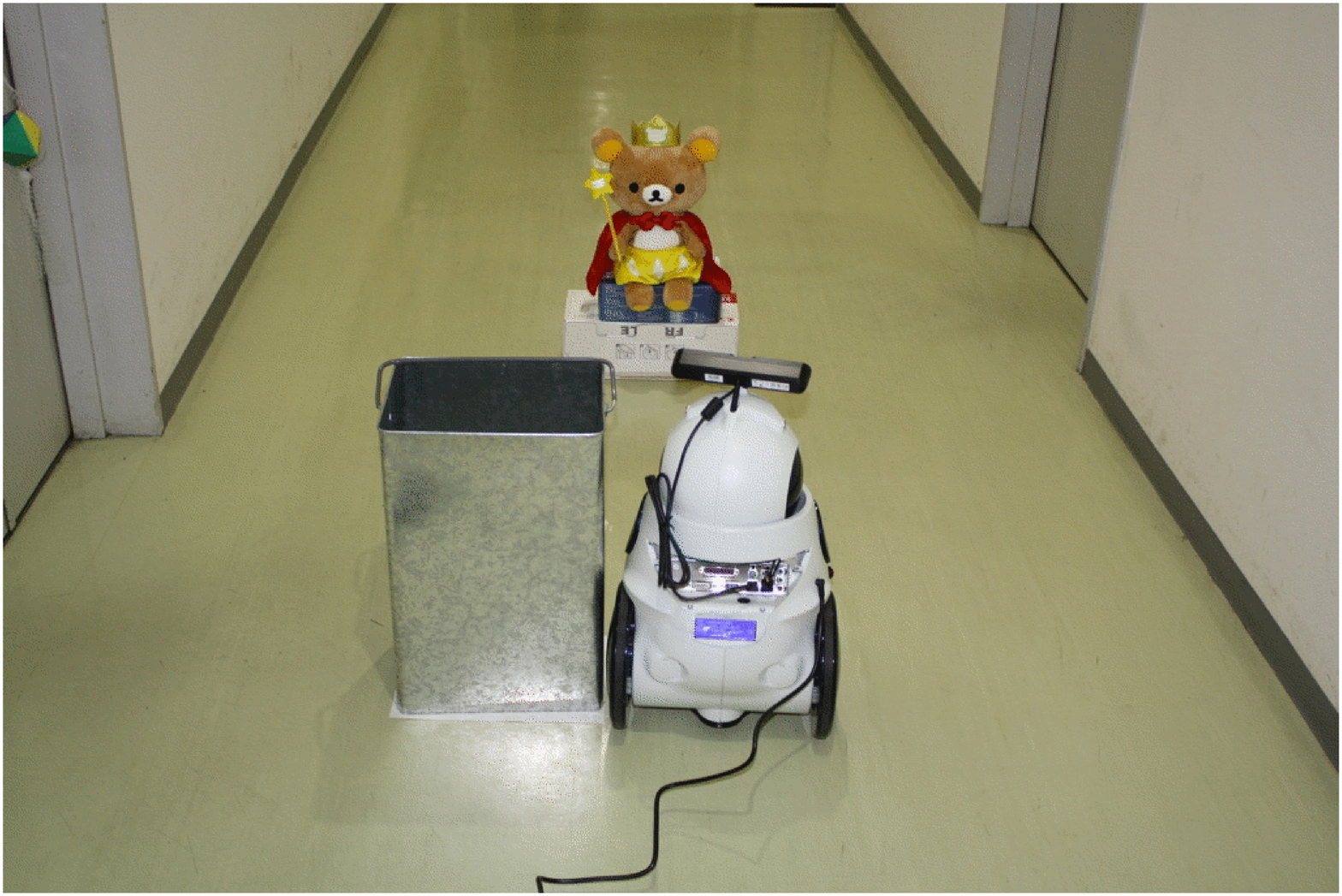}\\
(5) \QBO{} avoided obstacle.&(6) \QBO{} found target again.\\[\figsep]
\includegraphics[width=\figwidth]{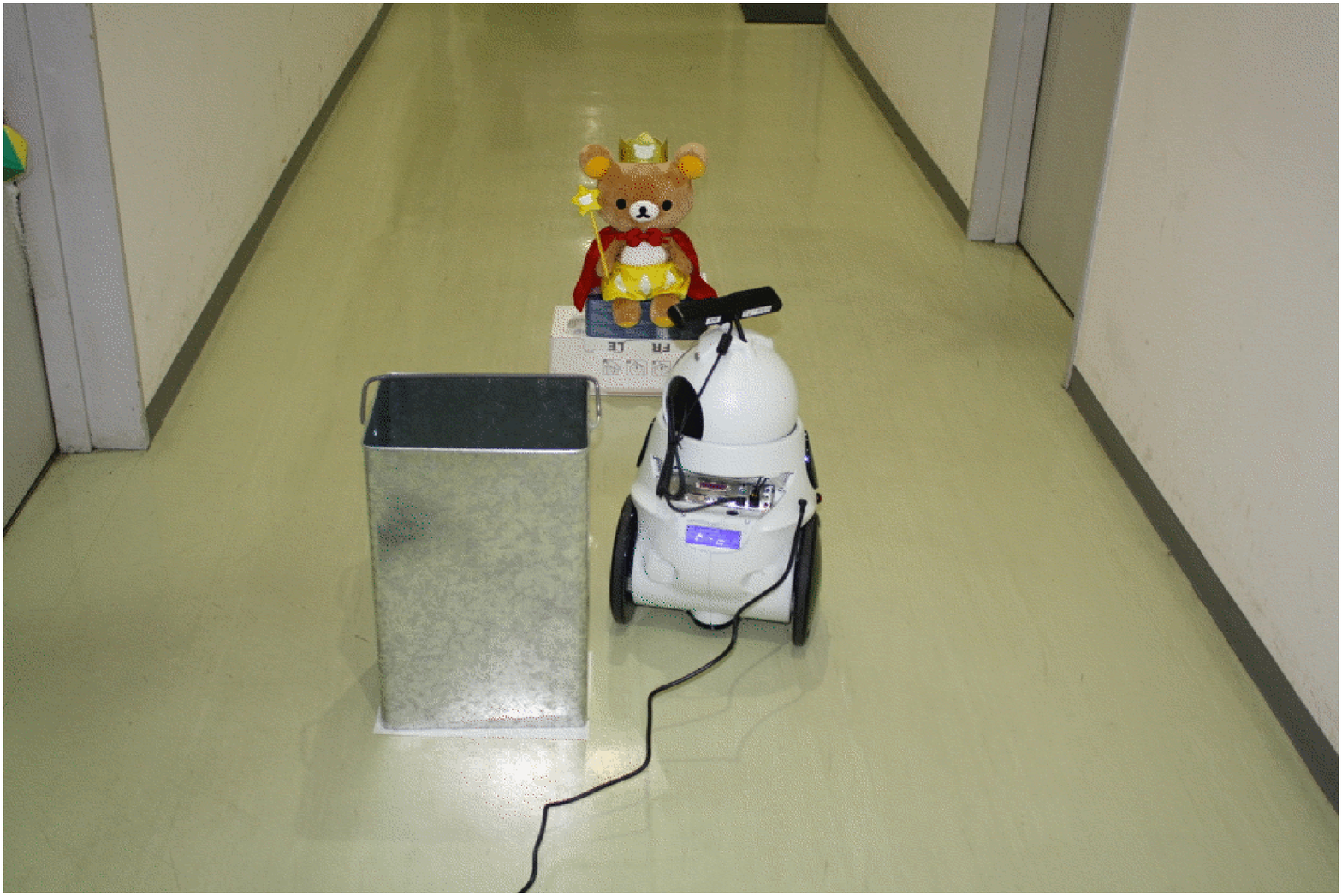}&
\includegraphics[width=\figwidth]{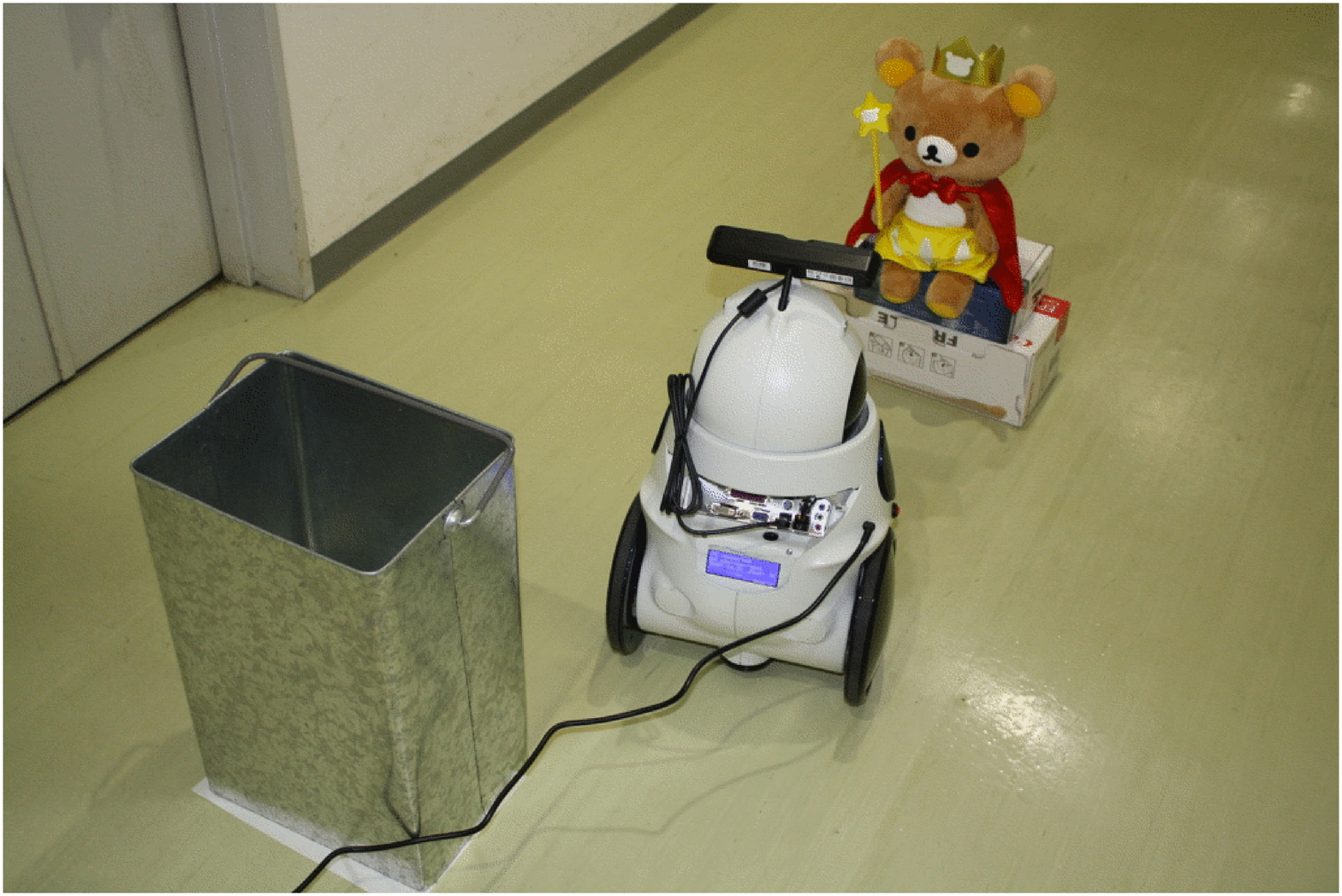}\\
(7) \QBO{} went towards target.&(8) \QBO{} reached a point
			in front of target.
\end{tabular}%
}
\caption{State of second experiment using \protect\QBO}
\label{Fig:experiment2}
\end{figure*}

The first experiment result is shown in \figref{Fig:experiment1}, and
the log messages of this experiment are shown in \figref{Fig:log}%
\footnote{In \figref{Fig:log}, \QBO's current direction and X, Y
coordinates are automatically calculated by ROS. However,
the X, Y coordinates are currently not used.}. The
result of the second experiment is shown
in \figref{Fig:experiment2}. We now give a detailed description of our
experiments of these figures in turn.

First, \QBO{}, the target and the obstacle are set on the floor as shown in
(1) of \figref{Fig:experiment1}. He found the target and started to
move forward to it (2). However, he could not go forward because of an
obstacle in front of him (3). He thus avoided this obstacle (4). Finally,
he reached a point in front of the target (5).

We also explain log messages \figref{Fig:log} in detail. He found the
target and the obstacle simultaneously ((1) and (2)). The central part
of \figref{Fig:log} shows the situation of (3) and (4). The end of this
shows the situation of (5).

In the second experiment, \QBO{} and the target were set on the floor ((1) as
shown in \figref{Fig:experiment2}). He found the target and went
towards it (2), and we put a tall obstacle in front of the target to
prevent him from seeing the target (3). Though he became unable to find the
direction of the target, he knew the direction which he faced at the
initial state. Thus, to continue along that direction he
avoided this obstacle (5) and he found the target again (6). He went
forward to the target (7), and he reached a point in front of
it (8).

\section{Related works}
\label{related_works}
Our method of generating and executing actions is
described in \secref{interleave}, is
logically much similar to that of
teleoreactive logic programs \cite{Langley06} in that
problem solving (action selection)
and skill execution are run in turn.
However,
most studies using teleoreactive logic programs
do not use actions associated with sensing;
they only deal with problems in which
the effects of actions can be accurately modeled
(\eg{} block worlds, driving in simulators).
In contrast, in our method,
the robot creates its behavior by concatenating
actions with sensing, such as `proceed until some obstacle is found'.
Thus, in the real world,
we can leave absorption of inaccuracy of actions
to low-level actions, and
high-level \decide{} can concentrate on
essential problem solving (such as route finding)\footnote{One possible
policy for classifying low- and high-level actions is to distinguish
reflexive actions and deliberative ones.
For example, when we go somewhere by bicycle, 
we keep our posture reflexively, and select a route to the
destination by deliberation.}.

On the other hand,
since the basis of making behaviors are common to both,
it is promising that
some techniques effective in teleoreactive logic programs
can also be applied to our method.
For example, \cite{Langley06} refers to learning skills
by generating logic programs. It is also possible in our
method in principle by dynamically asserting new Prolog clauses.

Several other pieces of work on robot control combine high-level goal
selection using Prolog and low-level controlled \kihons{}.

Pozo et al.'s{} \cite{Pozos07} presented a system which consists of a
planner based on the situation calculus written in Prolog and a Visual Basic
program to control a mobile robot via the serial port using the plan.
However, unlike our system, it is
based on common planning theory and does not take environmental
changes while performing \kihons{} into consideration.

Nalepa et al.{} \cite{Nalepa06} proposed the use of a Prolog-based
design of implementing embedded control systems.
However, it is principally aimed
at controlling devices such as mobile phones and elevator systems, and
does not take real-world robots into account. In a similar way, Matyasik
et al.'s work \cite{Matyasik07} provides a
controller of the Stenzel Ltd.{} HEXOR mobile robot, but it does not deal with
inaccuracies of robots' moves in the real world and lacks a way
to hide these inaccuracies from high-level action controls.

Qureshi et al.{} \cite{Qureshi08} provided a space robotics system by using a
combination of a visual perception system and a high-level reasoning system
using GOLOG \cite{Levesque97}.
It is specialized in space robotics tasks such as
rendezvous and docking while our system is intended to provide a
generalized architecture that combines high-level goal selections and
\kihons{} which are robust enough to bear up under inaccuracies in
the real world.

Some other researches introduce
actions associated with sensing, as we also do so.
A work of Chen et al.{} \cite{Chen10},
which intends human-robot collaboration (unlike ours),
can treat rather large-scale actions (\eg{}
`goto a location' and `pick-up an item') as atomic actions,
which can include sensing.
However, it does not directly mention the issues in the real world
described in \secref{real_world_fluctuation},
such as I/O disturbances, and failures in actions caused by them.
Besides, we think that `goto a location' is too large to be treated as
an atomic action; it should be better to regard it as a subgoal   
for which we can choose a plan by deliberation.

\textsc{KnowRob} by Tenorth et al.{} \cite{Tenorth09}
is an action-centered knowledge representation system
which can learn action models. It can build complex actions
including sensing as action classes.
It also mentions handling of uncertainties such as sensor noise.
However, since it is a knowledge representation system,
It does not mention how to overcome such uncertainties in itself.

\section{Conclusions}
\label{conclusion}
We present an architecture for motion planning, which can respond to
dynamic changes in the environment and also is able to
deal with uncertainties in the real world.
We incorporate sensing into \kihons{} because we
have to take the robot's embodiment into consideration.
\Kihons{} are similar to everyday actions done using sense data.

Though our experiment is still in the basic stage
and we are dealing with a small goal of reaching the single target,
we expect that our method will be able to handle
larger scale problem solving.
We plan to clarify the practicality of our method
through more pragmatic problems
such as handling multiple goals
and utilizing beliefs gained by past perceptions.
Increasing the precision of \kihons{} is
another future issue\footnote{As mentioned in \secref{nnaccurate},
it is a feature of our method that
the atomic actions need not be accurate enough.
However, increasing the precision of them can improve
the behaviors of robots.}.

While there are many works about planning, not many reports about
planning have been published
that deal with difficulties which come from embodiment
such as mentioned in \secref{real_world_fluctuation}.
We hope that this kind of
research increases as a result of ours.

\section*{Acknowledgments}
This work was supported by Nara Women's University
Intramural Grant for Young Women Researchers.

\bibliographystyle{splncs}
\bibliography{ref}

\end{document}